\documentclass[10pt, a4paper]{article}

\usepackage[final]{lrec-coling2024} 

\usepackage{booktabs}
\usepackage{multirow}
\usepackage{tabularx}
\usepackage{graphicx}
\usepackage{kotex}
\usepackage[super]{nth}
\usepackage[normalem]{ulem}
\usepackage{enumitem}
\useunder{\uline}{\ul}{}

\titlespacing{\paragraph}{0pt}{0.8\baselineskip}{1em}

\title{RECIPE4U: Student-ChatGPT Interaction Dataset \\in EFL Writing Education}

\name{Jieun Han$^*$, Haneul Yoo$^*$, Junho Myung, Minsun Kim, \\
    {\bf \large Tak Yeon Lee, So-Yeon Ahn, Alice Oh}} 

\address{KAIST \\
         South Korea \\
        \texttt{\{\href{mailto:jieun_han@kaist.ac.kr}{\color{black}{jieun\_han}}, \href{mailto:haneul.yoo@kaist.ac.kr}{\color{black}{haneul.yoo}}, \href{mailto:junho00211@kaist.ac.kr}{\color{black}{junho00211}}, \href{mailto:9909cindy@kaist.ac.kr}{\color{black}{9909cindy}}, }\\
        \texttt{
        \href{mailto:takyeonlee@kaist.ac.kr}{\color{black}{takyeonlee}},
        \href{mailto:ahnsoyeon@kaist.ac.kr}{\color{black}{ahnsoyeon}}\}@kaist.ac.kr},
        \texttt{alice.oh@kaist.edu}
}

\abstract{
    The integration of generative AI in education is expanding, yet empirical analyses of large-scale and real-world interactions between students and AI systems still remain limited.
Addressing this gap, we present RECIPE4U (RECIPE for University), a dataset sourced from a semester-long experiment with 212 college students in English as Foreign Language (EFL) writing courses.
During the study, students engaged in dialogues with ChatGPT to revise their essays.
RECIPE4U includes comprehensive records of these interactions, including conversation logs, students' intent, students' self-rated satisfaction, and students' essay edit histories. In particular, we annotate the students' utterances in RECIPE4U with 13 intention labels based on our coding schemes.
We establish baseline results for two subtasks in task-oriented dialogue systems within educational contexts: intent detection and satisfaction estimation.
As a foundational step, we explore student-ChatGPT interaction patterns through RECIPE4U and analyze them by focusing on students' dialogue, essay data statistics, and students' essay edits.
We further illustrate potential applications of RECIPE4U dataset for enhancing the incorporation of LLMs in educational frameworks.
RECIPE4U is publicly available at \url{https://zeunie.github.io/RECIPE4U/}. \\
 }

\begin{document}

\maketitleabstract

\newcommand{\edited}[1]{{\color{blue}{#1}}}

\section{Introduction}
The adoption of LLMs in education is accelerating, particularly in English as a Foreign Language (EFL) contexts \cite{han-etal-2023-recipe}. A noteworthy example is ChatGPT\thinspace\footnote{\label{chatgpt}\url{https://chat.openai.com/}}, a large language model (LLM)-driven chatbot developed by OpenAI, increasingly perceived as a beneficial resource for higher education students in research and writing tasks \cite{kasneci-2023-chatgpt}. EFL learners frequently face apprehension in exposing their linguistic shortcomings to instructors or peers \cite{cheng-2004-efl}. In this light, LLM-assisted tools tailored for English writing can alleviate such embarrassment by providing non-judgmental feedback. These tools foster a more supportive learning environment, as neither social distance nor a power relationship was found between students and the AI tools \cite{sun-2022-effects}. Nevertheless, most LLM-based tools, including ChatGPT, are not originally designed for educational purposes. 
As their application in EFL education grows, it is necessary to explore both the potential and the actual utilization patterns of LLMs in EFL writing education. Despite such needs, previous research falls short of conducting comprehensive, long-term analyses of real-world usage of LLMs within educational settings \cite {qadir-2023-engineering, baidoo-2023-education}.
RECIPE \cite{han-etal-2023-recipe} is the first attempt to suggest a platform to capture the semester-long interaction between students and ChatGPT in real-world EFL writing education. 

\begin{figure}
    \centering
    \includegraphics[width=\columnwidth]{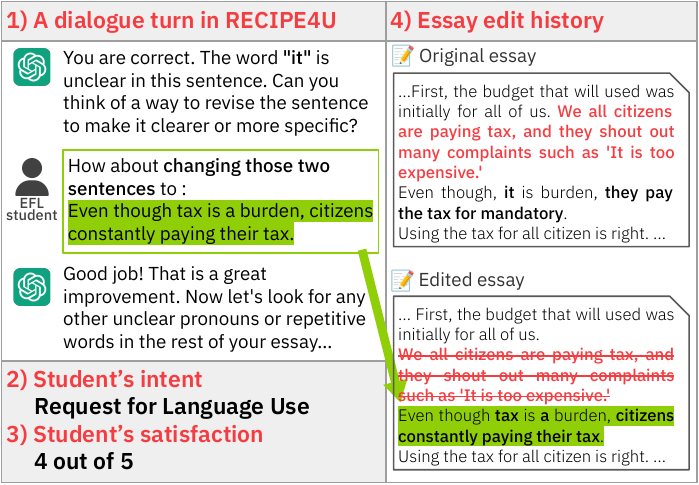}
    \vspace*{-7mm}
    \caption{Overview of RECIPE4U, a task-oriented dialogue dataset between EFL student and ChatGPT for essay revision. RECIPE4U includes 1) conversation log, 2) student's intent, 3) student's satisfaction, and 4) utterance-level student's essay edit history}
    \label{fig:teaser_image}
\end{figure}

We release RECIPE4U (RECIPE for University) dataset, which is derived from EFL learners in the university collected through the RECIPE platform. This dataset allows the detection of students' intent in their prompts (intent detection) and the estimation of their satisfaction with ChatGPT responses (satisfaction estimation), thereby establishing baseline models for two subtasks. We conduct an analysis of students' interaction patterns, contributing to a deeper understanding of the potential for future development in LLM-integrated English writing education.

The main contributions of this work include
\begin{enumerate}[itemsep=1pt]
    \item RECIPE4U (RECIPE for University), student-ChatGPT interaction dataset that captures semester-long learning process within the context of real-world EFL writing education.
    \item Baseline models for two subtasks with RECIPE4U: intent detection and satisfaction estimation.
    \item An investigation into the students' interaction patterns with conversation logs and essay edits in RECIPE4U for enhancing LLM-integrated education.
\end{enumerate}

\vspace{-2mm}
\section{Related Work}
\subsection{LLM-integrated Education}
There is a growing body of research exploring applications of LLM in educational contexts \cite{lee-2023-learning, markel-2023-gpteach, lu-2023-readingquizmaker}, but much of this research has focused on the short-term effects involving just a few experimental sessions. However, what we need is an investigation into long-term trends and usage patterns among students in the context of discourse analysis in education \cite{boyd-2015-relations} with a long-term analysis examining previous related interactions and contextual knowledge shared by students \cite{mercer-2008-seeds}. Also, LLM-integrated education is underexplored within the context of English as a foreign language (EFL) education where prior research \cite{qadir-2023-engineering, baidoo-2023-education, whalen-2023-chatgpt} has predominantly relied on episodic and anecdotal knowledge \cite{han-etal-2023-recipe}. In this paper, we investigate the potential of a systematic use of LLMs in EFL education in a semester-long university course.

\subsection{Dialogue Data in Education}
One of the common themes of educational dialogue analysis is the refinement and development of a coding scheme for dialogue acts \cite{ha-etal-2012-combining, boyer-etal-2010-dialogue, demszky-etal-2021-measuring, marineau-2000-classification}. However, previous work lacks annotation regarding students' underlying intentions or purpose behind their speech acts \cite{demszky-etal-2021-measuring, marineau-2000-classification}. 
\citet{demszky-etal-2021-measuring} identifies five uptake strategies related to teachers' utterances, which leaves out a comprehensive analysis of students' utterances. \citet{marineau-2000-classification} classifies the students' speech acts into four categories: assertions, wh-questions, yes/no questions, and directives. While these categories are useful for understanding the surface-level characteristics of students' utterances, they may not fully capture the nuances of their dialogue intentions.

Moreover, it is important to consider the specific domain of English education and the unique context of human-AI interaction. Currently available datasets in this field predominantly involve human-human interactions and do not examine the domain of EFL writing education \cite{demszky-etal-2021-measuring, rasor-2011-student}. To the best of our knowledge, our research represents a pioneering effort in introducing a dataset derived from real-world human-AI interactions within the context of EFL writing education. Additionally, we aim to shed light on the often-overlooked aspect of students' dialogue acts and their underlying intentions, thus contributing to a more comprehensive understanding of educational dialogues in this specific domain.

\subsection{Task-oriented Dialogue Dataset}
As a practical need in the industry, conversational AI has focused on delving into task-oriented dialogue (ToD) systems that can help specific daily-life tasks such as reservation and information query~\cite{hosseini-Asl-2020-simple}.
Various ToD datasets in the domain of everyday life were publicly released, including FRAMES~\citelanguageresource{el-asri-etal-2017-frames}, M2M~\citelanguageresource{shah2018building}, and DSTC-2~\citelanguageresource{henderson-etal-2014-second} for reservation and MultiWOZ 2.2~\citelanguageresource{zang-etal-2020-multiwoz}, KVRET~\citelanguageresource{eric-etal-2017-key}, SNIPS~\citelanguageresource{coucke2018snips}, and ATIS~\citelanguageresource{hemphill-etal-1990-atis} for information query, inter alia.
Recently, ToD systems also shed light on professional domains such as medical and education~\cite{wen-etal-2020-medal}.

To the best of our knowledge, \citetlanguageresource{zhang-etal-2023-groundialog} constructed GrounDialog, the first ToD dataset specifically tailored for language learning, with respect to repair and grounding (R\&G) patterns between high and low proficiency speakers of English.
However, GrouDialog deals with a task of information query on a job interview, which lacks relevance to language learning and education and is not publicly available.

Alongside the need for cross-lingual language modeling, several studies introduced multilingual ToD datasets, mostly constructed by translating English dialogues~\cite{schuster-etal-2019-cross-lingual, lin-2021-bitod}.
Still, code-mixed ToD datasets are hardly available~\cite{dowlagar-2023-code}.

\vspace{-2mm}
\section{RECIPE4U Dataset}

\begin{table*}[htb!]
\resizebox{\textwidth}{!}{
\begin{tabular}{@{}l|cccc|c@{}}
\toprule
                        & FRAMES                                                                                & M2M                                                                                 & MultiWOZ 2.2                                                              & GrounDialog                                                              & RECIPE4U (Ours)                                               \\ \midrule
\# of dialogues         & 1,369                                                                                 & 1,500                                                                               & 8,438                                                                     & 42                                                                       & 504                                                          \\
Total \# of utterances       & 19,986                                                                                & 14,796                                                                              & 115,424                                                              & 1,569                                                                    & 4,330                                                        \\
Total \# of tokens      & 251,867                                                                               & 121,977                                                                             & 1,520,970                                                                 & 14,566                                                                   & 380,364                                                      \\
Avg. utterances per dialogue & 14.60                                                                            & 9.86                                                                                & 13.68                                                                     & 37.36                                                                    & 3.38                                                         \\
Avg. tokens per utterance    & 12.60                                                                                 & 8.24                                                                                & 13.18                                                               & 9.28                                                                     & 87.84                                                         \\
Total unique tokens     & 12,043                                                                                & 1,008                                                                               & 24,071                                                                   & unknown                                                                  & 16,118                                                       \\
Languages               & English                                                                                    & English                                                                                  & English                                                                        & English                                                                       & English, Korean                                                       \\
Code-mixed              & X                                                                                     & X                                                                                   & X                                                                         & X                                                                        & O                                                            \\
Publicly available      & O                                                                                     & O                                                                                   & O                                                                         & X                                                                        & O                                                            \\
Additional data         & X                                                                                     & X                                                                                   & X                                                                         & X                                                                        & Essay edit history                                                            \\
Task                    & \begin{tabular}[c]{@{}c@{}}Find the best \\ deal of hotels \\ and flight\end{tabular} & \begin{tabular}[c]{@{}c@{}}Buy movie \\ ticket / Reserve\\  restaurant\end{tabular} & \begin{tabular}[c]{@{}c@{}}Get info. about \\ touristic city\end{tabular} & \begin{tabular}[c]{@{}c@{}}Get info. about \\ job interview\end{tabular} & \begin{tabular}[c]{@{}c@{}}Revise English \\ writing\end{tabular} \\ \bottomrule

\end{tabular}
}
\vspace{-3mm}
\caption{Statistics of RECIPE4U compared to existing task-oriented dialogue datasets}
\label{tab:tod_data_stats}
\end{table*}

We gather student-ChatGPT interaction data through RECIPE (Revising an Essay with ChatGPT on an Interactive Platform for EFL learners)~\cite{han-etal-2023-recipe}, a platform designed to integrate ChatGPT with essay writing for EFL students. The main component of RECIPE is a writing exercise where students write and revise their essays while conversing with ChatGPT. We provide students with instructions to revise an essay while having a conversation about what they learned in class. The student is shown a user interface that AI agent initiates the conversation by requesting a class summary. RECIPE incorporates \texttt{gpt-3.5-turbo} prompted with 1) a persona of an English writing class teacher and 2) step-by-step guidance to students in the platform. 

A semester-long longitudinal data collection involves 212 EFL students (91 undergraduate and 121 graduate students) from a college in South Korea. They are enrolled in one of the three different English writing courses: Intermediate Writing, Advanced Writing, and Scientific Writing. Undergraduate students were divided into two courses depending on their TOEFL writing scores (15-18 for Intermediate Writing and 19-21 for Advanced Writing). In both courses, one of the primary assignments is writing an argumentative essay. Scientific Writing course is designed for graduate students, aiming to teach them how to write scientific research papers. 

In total, RECIPE4U contains 4330 utterances (1913 students' utterances and 2417 ChatGPT's utterances), including 97 single-turn and 407 multi-turn dialogues. The conversation is mostly done in English, but there are several instances of code-switching between English and Korean, as the majority of the student's first language is Korean. 
Table~\ref{tab:tod_data_stats} describes detailed statistics of RECIPE4U dataset compared to existing task-oriented dialogue datasets.

Unlike other task-oriented dialogue dataset, RECIPE4U includes additional source data, which is 1913 utterance-level essay edit history.
We collect students' essay edit history at each utterance level to explore students' learning process. 
Students voluntarily provide their essays to RECIPE~\cite{han-etal-2023-recipe} and make necessary edits while having a conversation with ChatGPT on topics regarding essay writing.

We gather students' self-rated satisfaction levels to analyze the students' learning experiences and gain insights into how they perceived and evaluated their interactions with ChatGPT on a five-Likert scale. Specifically, each time a student engages in a conversation, RECIPE asks students to self-rate their level of satisfaction with ChatGPT's last response. In addition, we add tags to the students' written essays, paragraphs, and sentences for future application.

\section{Experiment}
\begin{table*}[htb!]
\centering
\resizebox{\linewidth}{!}{
\begin{tabular}{@{}m{0.21\textwidth}|m{0.3\textwidth}|m{0.35\textwidth}|r@{}}
\toprule
Intent                                                             & Definition                                                                                                       & Example                                                                                                                              & Distribution \\ \midrule
Acknowledgement                                                    & The student acknowledges previous utterance; conversational grounding                                            & \textit{\begin{tabular}[c]{@{}l@{}}AI: Please provide your essay.\\ Student: {[}ESSAY{]}\end{tabular}}                               & 16.52\%    \\ \midrule
Negotiation                                                        & The student negotiates with the teacher on shared activity                                                       & \textit{The first two mistakes that you pointed out make no sense
}                                                                & 5.07\%     \\ \midrule
Other                                                              & Other utterances, usually containing only affective content                                                      & \textit{Hi there}                                                                                                                    & 1.25\%     \\ \midrule
\begin{tabular}[c]{@{}m{0.21\textwidth}@{}}Request for\\ Translation\end{tabular}  & The student requests for translation on the utterance of either himself/herself or AI.                           & \textit{Can you translate it into Korean?
}                                                                & 1.25\%     \\ \midrule
\begin{tabular}[c]{@{}m{0.21\textwidth}@{}}Request for\\ Confirmation\end{tabular} & The student requests confirmation from the teacher                                                               & \textit{I bet there shouldn't be a lot of citation appear in the introduction, should it?}                          & 1.15\%     \\ \midrule
\begin{tabular}[c]{@{}m{0.21\textwidth}@{}}Request for\\ Language Use\end{tabular} & The student requests for evaluation or revision or information on grammar, vocabulary, spelling, and punctuation & \textit{Please point out the grammatical mistakes in the essay}                                                    & 16.00\%    \\ \midrule
\begin{tabular}[c]{@{}m{0.21\textwidth}@{}}Request for\\ Revision\end{tabular}     & The student requests revision from the tutor other than language use                                             & \textit{Can you change this paragraph to be more effective to read?}                                                                 & 8.57\%     \\ \midrule
\begin{tabular}[c]{@{}m{0.21\textwidth}@{}}Request for\\ Evaluation\end{tabular}   & The student requests an evaluation from the tutor other than language use                                        & \textit{Could you check if my essay has unity and coherence? Here is my essay.}                                                      & 7.11\%     \\ \midrule
\begin{tabular}[c]{@{}m{0.21\textwidth}@{}}Request for\\ Information\end{tabular}  & The student requests information from the tutor other than language use                                          & \textit{What are the common mistakes people do when writing an abstract?}                                                            & 13.12\%    \\ \midrule
\begin{tabular}[c]{@{}m{0.21\textwidth}@{}}Request for\\ Generation\end{tabular}   & The student requests generation from the tutor other than language use                                           & \textit{Could you write it for me?}                                                                                                  & 3.14\%     \\ \midrule
Question                                                           & A question regarding the task that is not a request for confirmation or feedback or translation or language use  & \textit{Ahh... How many neurons do you have? I'm so curious about your structure and size.}                                          & 2.88\%     \\ \midrule
Answer                                                             & An answer to an utterance to request or question from the tutor                                                  & \textit{\begin{tabular}[c]{@{}m{0.35\textwidth}@{}}AI: Can you please tell me what you learned in class? \\ Student: Today I learned….\end{tabular}} & 21.01\%    \\ \midrule
Statement                                                          & A statement regarding the task that does not fit into any of the above categories                                & \textit{Also believe, think}                                                                                                                            & 2.93\%     \\ \bottomrule
\end{tabular}
}
\vspace{-3mm}
\caption{Definition and sample utterances for student dialogue intent}
\label{tab:intent_label}
\end{table*}

\begin{table*}[htb!]
\centering
\resizebox{0.85\textwidth}{!}{
\begin{tabular}{@{}ll|rrrrrr|c@{}}
\toprule
\multicolumn{2}{c|}{Intent}                           & \multicolumn{6}{c|}{Satisfaction} & \multirow{2}{*}{\begin{tabular}[c]{@{}c@{}}Avg. of\\ Satisfaction\end{tabular}} \\
\multicolumn{1}{c}{div1}  & \multicolumn{1}{c|}{div2} & 1  & 2  & 3   & 4   & 5   & Total &                                                                                 \\ \midrule
\multirow{3}{*}{Response} & Acknowledgement           & 2  & 10 & 21  & 106 & 177 & 316   & 4.41                                                                           \\
                          & Negotiation               & 4  & 8  & 18  & 32  & 35  & 97    & 3.89                                                                           \\
                          & Answer                    & 8  & 9  & 63  & 165 & 157 & 402   & 3.83                                                                           \\ \midrule
\multirow{8}{*}{Request}  & Request for Translation   & 2  & 2  & 4   & 9   & 7   & 24    & 3.71                                                                           \\
                          & Request for Confirmation  & 1  & 1  & 3   & 10  & 7   & 22    & 3.96                                                                           \\
                          & Request for Language Use  & 7  & 13 & 42  & 114 & 130 & 306   & 4.13                                                                           \\
                          & Request for Revision      & 6  & 6  & 28  & 58  & 66  & 164   & 4.05                                                                           \\
                          & Request for Evaluation    & 8  & 8  & 17  & 50  & 53  & 136   & 3.97                                                                           \\
                          & Request for Information   & 6  & 10 & 28  & 97  & 110 & 251   & 4.18                                                                           \\
                          & Request for Generation    & 1  & 7  & 5   & 21  & 26  & 60    & 4.07                                                                           \\
                          & Question                  & 4  & 6  & 14  & 15  & 16  & 55    & 3.60                                                                           \\ \midrule
\multirow{2}{*}{Misc.}    & Statement                 & 2  & 1  & 8   & 19  & 26  & 56    & 4.13                                                                           \\
                          & Other                     & 4  & 0  & 4   & 4   & 12  & 24    & 4.18                                                                           \\ \midrule
Total                     &                           & 55 & 81 & 255 & 700 & 822 & 1913  & 4.13                                                                           \\ \bottomrule
\end{tabular}
}
\vspace{-1mm}
\caption{Number of samples by intent and satisfaction}
\label{tab:num_of_samples}
\end{table*}

In this paper, we suggest two subtasks leveraging RECIPE4U data: intent detection (\S\ref{sec:intent_detection}) and satisfaction estimation (\S\ref{sec:satisfaction_estimation}). 

\vspace{-2mm}
\subsection{Intent Detection}
\label{sec:intent_detection}
Intent detection is the task of classifying the students' utterances into 13 predefined intent categories. This task involves intent classification based on a student's utterance, the preceding response from ChatGPT, and the subsequent response from ChatGPT. 

\vspace{-2mm}
\subsubsection{Intent Label}
We design students' intention annotation schemes, comprising 13 intention labels. This scheme builds upon and complements the analysis of intention in dialogue from previous research \cite{ha-etal-2012-combining, boyer-etal-2010-dialogue, ozkose-2015-plays}. From the study on task-oriented dialogue in educational settings by \citet{ha-etal-2012-combining}, we adopted eight intention labels: \textsc{Acknowledgement}, \textsc{Other}, \textsc{Request for Confirmation}, \textsc{Request for Revision}, \textsc{Request for Evaluation}, \textsc{Question}, \textsc{Answer}, and \textsc{Statement}. From \citet{ozkose-2015-plays}'s research on learner reciprocity, we integrated the \textsc{Negotiation} and \textsc{Request for Information} labels. To better cater to the context of student-AI interactions in EFL writing education, we introduced three additional labels: \textsc{Request for Translation}, \textsc{Request for Language Use}, and \textsc{Request for Generation}. These 13 intention labels can be further grouped into three ancestor categories, division 1 (div1): \textsc{Response}, \textsc{Request}, and \textsc{Miscellaneous}.
Under the \textsc{Response} category, we include three labels from division 2 (div2): \textsc{Acknowledgement}, \textsc{Negotiation}, and \textsc{Answer}. The \textsc{Request} category encompasses eight labels: \textsc{Request for Translation}, \textsc{Request for Confirmation}, \textsc{Request for Language Use}, \textsc{Request for Revision}, \textsc{Request for Evaluation}, \textsc{Request for Information}, \textsc{Request for Generation}, and \textsc{Question}. Lastly, the \textsc{Miscellaneous} class includes \textsc{Statement} and \textsc{Other}. The descriptions and examples of 13 labels are shown in Table~\ref{tab:intent_label}. 

\vspace{-2mm}
\subsubsection{Intent Annotation}
Four authors engage in an iterative process of collaborative and independent tagging of the student's intent. In the initial tagging phase, all annotators collaboratively tag 10.45\% of the dialogue dataset. Subsequently, the remaining samples undergo annotation independently by two annotators. 
In cases where disagreements arose, all four authors again discuss once more to tag until a consensus is reached.

Table~\ref{tab:num_of_samples} shows the distribution of student intentions and satisfaction levels. The most frequent intention was \textsc{Answer}, followed by \textsc{Acknowledgement}, and \textsc{Request for Language Use}. The top two labels suggest a high level of compliance and engagement among students when interacting with ChatGPT. Also, it is notable that EFL learners primarily utilize ChatGPT to seek assistance with language use. The low frequency of \textsc{Request for Confirmation} and \textsc{Other} aligns with the findings from previous work that analyzed the student-human tutor dialogue acts in real-world tutoring sessions \cite{ha-etal-2012-combining}.

\subsection{Satisfaction Estimation}
\label{sec:satisfaction_estimation}
Satisfaction estimation is a classification task to predict the students' satisfaction with the ChatGPT's last response. This estimation was conducted on a turn-level, leveraging the users' self-ratings collected from RECIPE as a gold label. The task is done in two distinct settings: binary classification and ordinal classification. Binary classification involves an estimation of whether the given utterance is helpful or not. In this context, a satisfaction score falling within the range of 1 to 2 was considered an unhelpful utterance, while a score in the range of 4 to 5 was considered helpful. For ordinal classification, models were tested for their ability to estimate the students' exact satisfaction score on a scale of 1 to 5.

\subsection{Experimental Results}
We use multilingual BERT~\cite{devlin-etal-2019-bert}, XLM-R~\cite{conneau-etal-2020-unsupervised}, \texttt{gpt-3.5-turbo-16k} and \texttt{gpt-4}\thinspace\footnote{The experiments were conducted on September 30, 2023 - October 2, 2023.} for the experiment. 
We choose multilingual models that support both English and Korean, considering code-switched utterances in RECIPE4U.
As input text for intent detection, we use a set of previous responses of ChatGPT, user utterances, and subsequent responses of ChatGPT, and for satisfaction estimation, we use a pair of user utterances and subsequent responses, respectively, without any essay component tags.
We examine fine-tuned M-BERT~\cite{devlin-etal-2019-bert} and XLM-R~\cite{conneau-etal-2020-unsupervised} with 5-fold validations and infer \texttt{gpt-3.5-turbo-16k} and \texttt{gpt-4} with five different prompts. We experiment \texttt{gpt-3.5-turbo-16k} and \texttt{gpt-4} under four different settings with 0.2 and 1.0 temperature (temp.) and with zero and few-shot.  
We use Intel(R) Xeon(R) Silver 4114 (40 CPU cores) and GeForce RTX 2080 Ti 10GB (4 GPUs) for fine-tuning M-BERT~\cite{devlin-etal-2019-bert} and XLM-R~\cite{conneau-etal-2020-unsupervised}.

\begin{table*}[th!]
\centering
\resizebox{\linewidth}{!}{
\begin{tabular}{@{}lcc|cc|cc@{}}
\toprule
                                            &             &            & \multicolumn{2}{c|} {Intent Detection}  & \multicolumn{2}{c}{Satisfaction Estimation} \\
                                            & temp.        & shot       & div1 (3 cls)                     & div2 (13 cls)                    & div1 (2 cls)                     & div2 (5 cls)                     \\ \midrule
M-BERT~\cite{devlin-etal-2019-bert}         & \multicolumn{2}{c|}{N/A} & \textbf{0.8344}$_{\pm0.0494}$    & \underline{0.4291}$_{\pm0.0330}$    & \textbf{0.9109}$_{\pm0.0095}$    & \textbf{0.5794}$_{\pm0.1025}$    \\
XLM-R~\cite{conneau-etal-2020-unsupervised} & \multicolumn{2}{c|}{N/A} & 0.7041$_{\pm0.0291}$             & 0.2849$_{\pm0.0530}$             & 0.8591$_{\pm0.0135}$             & 0.4436$_{\pm0.0896}$             \\ \midrule
\multirow{4}{*}{\texttt{gpt-3.5-turbo-16k}} & 0.2         & zero       & 0.1585$_{\pm0.0085}$             & 0.2261$_{\pm0.0066}$             & \underline{0.8745}$_{\pm0.0110}$ & 0.4886$_{\pm0.0446}$             \\
                                            & 1.0         & zero       & 0.1581$_{\pm0.0040}$             & 0.2251$_{\pm0.0066}$             & 0.8384$_{\pm0.0272}$             & 0.4604$_{\pm0.0352}$             \\
                                            & 0.2         & few        & 0.6202$_{\pm0.0243}$             & 0.3053$_{\pm0.0370}$             & 0.7551$_{\pm0.0173}$             & 0.4435$_{\pm0.0179}$             \\
                                            & 1.0         & few        & 0.5261$_{\pm0.0146}$             & 0.2047$_{\pm0.0229}$             & 0.7322$_{\pm0.0220}$             & 0.3968$_{\pm0.0165}$             \\ \midrule
\multirow{4}{*}{\texttt{gpt-4}}             & 0.2         & zero       & \underline{0.7779}$_{\pm0.0167}$ & \textbf{0.4891}$_{\pm0.0203}$ & 0.8626$_{\pm0.0039}$             & \underline{0.5639}$_{\pm0.0133}$ \\
                                            & 1.0         & zero       & 0.7669$_{\pm0.0153}$             & 0.4763$_{\pm0.0210}$             & 0.8582$_{\pm0.0046}$             & 0.5469$_{\pm0.0130}$             \\
                                            & 0.2         & few        & 0.6991$_{\pm0.0776}$             & 0.4661$_{\pm0.0568}$             & 0.8188$_{\pm0.0080}$             & 0.4768$_{\pm0.0051}$             \\
                                            & 1.0         & few        & 0.6678$_{\pm0.0645}$             & 0.4359$_{\pm0.0538}$             & 0.8069$_{\pm0.0075}$             & 0.4618$_{\pm0.0089}$             \\ \bottomrule
\end{tabular}
}
\vspace{-3mm}
\caption{Experimental results (micro-averaged F1 scores) for intent detection and satisfaction estimation}
\label{tab:experimental_result}
\vspace{+2mm}
\end{table*}
Table ~\ref{tab:experimental_result} shows experimental results measured by micro-averaged F1 scores for intent detection and satisfaction estimation.
Fine-tuned BERT~\cite{devlin-etal-2019-bert} achieves the highest results across all tasks, followed by \texttt{gpt-4} with zero-shot and a temperature of 0.2, in general.

\subsection{Ablation Study}
In this section, we examine several conditions to boost model performances in our proposed tasks.
We conduct all ablation study experiments using the fine-tuned BERT~\cite{devlin-etal-2019-bert}, which outperforms other models as shown in Table~\ref{tab:experimental_result}.

\paragraph{Whose utterance is critical?}
We investigate the impact of various utterances --- prior responses from ChatGPT, user utterances, and subsequent responses from ChatGPT --- on our tasks.
The findings are presented in Table~\ref{tab:ablation_input}.
Combining the user's utterance with the subsequent response from ChatGPT as a pair offers the best results, closely followed by using the user's utterance only.
\begin{table*}[htb!]
\centering
\resizebox{0.95\linewidth}{!}{
\begin{tabular}{@{}ccc|cc|cc@{}}
\toprule
\multicolumn{3}{c|}{Input utterances}     & \multicolumn{2}{c|}{Intent Detection}                         & \multicolumn{2}{c}{Satisfaction Estimation}                   \\
ChatGPT$_{i-1}$ & User$_i$ & ChatGPT$_{i+1}$ & div1 (3 cls)                  & div2 (13 cls)                 & div1 (2 cls)                  & div2 (5 cls)               \\ \midrule
                 & O    &                 & {\ul 0.8327}$_{\pm0.0387}$    & {\ul 0.4366}$_{\pm0.0544}$    & {\ul 0.8812}$_{\pm0.0160}$     & \textbf{0.5878}$_{\pm0.0034}$ \\
O                & O    &                 & 0.8272$_{\pm0.0478}$          & 0.4301$_{\pm0.0545}$          & {\ul 0.8812}$_{\pm0.0160}$     & 0.5035$_{\pm0.1019}$          \\
                 & O    & O               & 0.8280$_{\pm0.0323}$          & \textbf{0.4942}$_{\pm0.0382}$ & \textbf{0.9109}$_{\pm0.0095}$ & {\ul 0.5794}$_{\pm0.1025}$    \\
O                & O    & O               & \textbf{0.8344}$_{\pm0.0494}$ & 0.4291$_{\pm0.0330}$          & {\ul 0.8812}$_{\pm0.0160}$     & 0.5787$_{\pm0.0276}$          \\ \bottomrule
\end{tabular}
}
\vspace{-2mm}
\caption{Ablation study on input utterances using fine-tuned BERT. $E_i$ denotes the $i$-th utterance spoken by the entity $E$.}
\label{tab:ablation_input}
\end{table*}

\paragraph{Should we mask essays in dialogue?}
\begin{table*}[htb!]
\centering
\resizebox{0.75\textwidth}{!}{
\begin{tabular}{@{}l|cc|cc@{}}
\toprule
\multicolumn{1}{c|}{\multirow{2}{*}{Input}} & \multicolumn{2}{c|}{Intent Detection}                         & \multicolumn{2}{c}{Satisfaction Estimation}                   \\
\multicolumn{1}{c|}{}                       & div1 (3 cls)                  & div2 (13 cls)                 & div1 (2 cls)                  & div2 (5 cls)                  \\ \midrule
raw texts                                   & {\ul 0.8344}$_{\pm0.0494}$    & 0.4291$_{\pm0.0330}$          & {\ul 0.9109}$_{\pm0.0095}$    & {\ul 0.5794}$_{\pm0.1025}$    \\
+ special token                             & 0.8279$_{\pm0.0257}$          & \textbf{0.4631}$_{\pm0.0472}$ & 0.9083$_{\pm0.0021}$          & 0.5527$_{\pm0.1120}$          \\
+ masking                                   & \textbf{0.8473}$_{\pm0.0487}$ & {\ul 0.4325}$_{\pm0.0797}$    & \textbf{0.9112}$_{\pm0.0096}$ & \textbf{0.6002}$_{\pm0.0661}$ \\ \bottomrule
\end{tabular}
}
\vspace{-3mm}
\caption{Ablation study on masking essays in dialogue using fine-tuned BERT}
\label{tab:ablation_essay}
\end{table*}
Since RECIPE4U is a task-oriented dialogue aiming to revise EFL students' essays, both utterances from the user and ChatGPT often incorporate entire essays or fragments thereof, including paragraphs and sentences. 
We evaluate how these essay components within dialogues affect our task predictions.
First, we introduce special tokens such as \texttt{<sentence>}, \texttt{<paragraph>}, and \texttt{<essay>} to delineate essay components.
We also mask essay components and replace them with special tokens.
As illustrated in Table~\ref{tab:ablation_essay}, performance generally peaks when masking essay components.
The inclusion of special tokens did not significantly differentiate from providing raw input. 
We hypothesize that masking essay components shortens the input texts, which mitigates the token limit issue and focuses on the core part of the utterances.

\section{Discussion}
We delve into students' interaction with ChatGPT through quantitative and qualitative analysis, focusing on 1) students' dialogue patterns, 2) essay data statistics, and 3) essay edit patterns. In the following section, we will use the notation S$n$ to represent individual students for sample-level analysis, with $n$ denoting the student sample ID.
\subsection{Students' Dialogue Patterns}
\label{sec:usage}
In our analysis of students' dialogues, we identify that students tend to perceive ChatGPT as a human-like AI, as a multilingual entity, and as an intelligent peer. 

\paragraph{As human-like AI} 
Despite being aware that they are conversing with ChatGPT, students often tend to anthropomorphize it. They frequently refer to ChatGPT by name and express gratitude towards it, suggesting that students perceived ChatGPT as possessing its own personality and emotions. 
This tendency towards anthropomorphism positively influences the quality of interaction and students' acceptance of AI \cite{pelau-2021-makes}. S1 addresses ChatGPT by name, saying \textit{``Hey ChatGPT you said\ldots''}. Furthermore, 113 samples included expressions of gratitude towards ChatGPT for its guidance. S2 and S3 both compliment ChatGPT and convey gratitude by saying \textit{``Wow great! Thank you so much''} and \textit{``yup that's perfect. thank you!''}  respectively. S4 even states \textit{``Thank you, ChatGPT''}, demonstrating both gratitude and recognition of its name.

\paragraph{As multilingual entity} 
Code-switching is a common phenomenon observed in the utterances of EFL learners, and they use it to express annoyance as well as respect \cite{silaban-2020-analysis}. Students' utterances in RECIPE4U also included code-switching, expecting ChatGPT to understand their requests in both languages. When dissatisfied with ChatGPT's response, students often switch to their first language, Korean. For instance, S5 initially inquire in English, asking, \textit{``Is 7 and 8 grammatical error?''} regarding the grammatical errors in the essay. After receiving a response from ChatGPT, S5 rates the satisfaction as 2 and then expressed doubt on ChatGPT's response by saying \textit{``7번 문장에서 other말고 다른 부분은 문법적 오류가 아니지 않아? (Isn’t the rest of the sentence except for ‘other’ in sentence 7 free of grammatical error?)''} in Korean, which is his or her first language. On the other hand, students code-switch to acknowledge ChatGPT's utterance. For example, S6 first asks ChatGPT in Korean, \textit{``좌측의 내 에세이에서 틀린점을 짚어줘 (Please point out the errors in my essay on the left.)''} As ChatGPT responds, \textit{``Could you please provide the instruction in English so that I can guide you through the revision process?''}, the student then acknowledges the request from ChatGPT by switching Korean to English, \textit{``Please point out the mistakes in my essay on the left.''}

\paragraph{As intelligent peer} 
Students often perceive ChatGPT as an approachable and intelligent peer rather than viewing it as an instructor. They feel comfortable asking questions to ChatGPT that they might not ask the professor. Students seek clarification from ChatGPT when they encounter concepts or feedback that they did not fully understand during lectures delivered by their professors. 
For instance, S5 and S6 challenge their professors' statements by saying, \textit{``that's what i selected too but my professor marked that `cannot' should also have been selected [\textit{sic}].''} and \textit{``But my Professor said it is healthful [\textit{sic}]''}.
However, the friendliness towards ChatGPT can lead to academic integrity problems, as students can simply ask ChatGPT anything whenever they want. Our analysis reveals 22 samples where students heavily rely on ChatGPT, asking ChtGPT to provide answers for quizzes and assignments instead of attempting to solve them independently.


\subsection{Students' Essay Statistics}
\begin{table}[htb!]
\resizebox{\columnwidth}{!}{
\begin{tabular}{@{}ll|cc@{}}
\toprule
                       &                       & First draft       & Final draft          \\ \midrule
\multirow{4}{*}{score} & content (/5)$^*$      & 3.34$_{\pm0.59}$    & 3.66$_{\pm0.57}$     \\
                       & organization (/5)$^*$ & 3.54$_{\pm0.79}$    & 3.80$_{\pm0.63}$      \\
                       & language (/5)$^*$     & 3.20$_{\pm0.59}$     & 3.71$_{\pm0.72}$     \\
                       & overall (/15)$^*$     & 10.08$_{\pm1.62}$   & 11.17$_{\pm1.54}$    \\ \midrule
\multirow{3}{*}{stats} & \# of tokens$^*$      & 314.13$_{\pm82.80}$ & 408.41$_{\pm187.87}$ \\
                       & \# of sentences$^*$   & 19.30$_{\pm5.93}$    & 23.27$_{\pm9.46}$    \\
                       & \# of paragraphs      & 4.09$_{\pm3.88}$    & 4.79$_{\pm5.10}$     \\ \midrule
\multicolumn{2}{l|}{perplexity$^*$}            & 38.57$_{\pm20.52}$  & 27.28$_{\pm14.23}$   \\ \bottomrule
\end{tabular}
}
\vspace{-3mm}
\caption{Statistics of essays in RECIPE4U dataset. The asterisk denotes a statistically significant difference tested by paired t-test ($p$-value < 0.01).}
\label{tab:essay_stats}
\end{table}
REICPE4U captures unique interactions of students throughout the semester, providing valuable insights into the students' learning processes. We measure students' learning process through semester-long interaction data in RECIPE4U.
Table~\ref{tab:essay_stats} depicts statistics of the first and the final essays from each student.
We evaluate essays with three standard EFL essay scoring rubrics: content, organization, and language~\cite{han-2023-fabric}, and calculate perplexity using GPT-2. 
We discover a notable improvement across all aspects of the students' essays.
Specifically, there are significant differences between the two drafts in terms of essay scores, token count, sentence count, and perplexity.

\vspace{-3mm}
\subsection{Students' Essay Edit Patterns}
Students' essay edit history in RECIPE4U can offer deeper insights, especially when examining their reception of feedback from ChatGPT. This information provides a lens into EFL students' behavior and perceptions towards ChatGPT's essay improvement suggestions.
\paragraph{Accepting ChatGPT feedback}
In the RECIPE4U dataset, there are 351 instances where students make edits to their essays during interactions with ChatGPT. The top three \textsc{Request} prompts that lead students to edit their essay after receiving the response from ChatGPT are \textsc{Request for Language Use}, \textsc{Request for Revision}, \textsc{Request for Information}. It suggests that EFL learners are particularly receptive to feedback concerning language usage. This inclination resonates with the observation that students often accept feedback from ChatGPT, especially when it addresses language errors. S7 accepted the feedback by deleting `more' before the comparative `less', stating to change \textit{``a more less passive life''} into \textit{``a less passive life''}.

In addition, some students directly request to write an essay rather than seeking guidance and simply copy-paste what ChatGPT generated. S8 inquires, \textit{``Could you write it for me?''} to ChatGPT. In response, ChatGPT generates a paragraph which S8 then copies into their essay without any revision. In contrast, there are cases where ChatGPT chooses not to fulfill such requests. When S9 requests \textit{``Then, could you rewrite my essay with these ideas?''}, ChatGPT does not complete the essay, underscoring its educational role by stating it should assist with the writing process.

\paragraph{Rejecting ChatGPT feedback}
The essay edits do not necessarily mean that students have embraced the feedback from ChatGPT. There are various reasons that might prompt students to disregard the suggestions, including feedback deemed trivial, unintended change, and ChatGPT's hallucination. 

Students might overlook feedback that they perceive as minor or trivial. This often encompasses recommendations to adjust slight expressions or grammatical elements. For example, when ChatGPT suggests a vocabulary edit of `largely' into `broadly', which entails a similar meaning, S10 chooses not to modify the essay. Likewise, S7 perceives an article error as less important and leaves it unchanged when ChatGPT correctly points out an article error of \textit{``Computer science student''} to \textit{``A computer science student''}.

ChatGPT often offers unsolicited changes, either editing sections not asked for or altering the essay's intended meaning.
In response to a request from S11 to identify typos, ChatGPT opted for a broader rephrase: from \textit{``they can choose and design their future on their own''} to \textit{``they can make informed and thoughtful decisions about their future''}. As another example, when S5 asks for only grammatical errors only, ChatGPT highlights spelling issues and phrasing nuances. This leads the student to further clarify his or her initial request, stating, \textit{``그럼 1번부터 10번까지 문법적 오류만 교정해서 다시 알려줘 (Then please tell me again after correcting only grammatical errors from 1 to 10)''}.


ChatGPT may produce feedback based on incorrect assumptions or inaccuracies. Such hallucinations from ChatGPT can lead to suggestions that aren't applicable to the student's actual essay. A case in point is when ChatGPT advises, \textit{``\ldots you need a comma after `violent' to separate two adjectives''}, but the word `violent' did not exist in the student's essay. This results in S12 querying ChatGPT, expressing their confusion with, \textit{``I don't know where I wrote the word `violent'''}.

\subsection{Future Direction}
\label{sec:future_guide}
In this section, we outline potential human-LLM collaboration approaches to further develop LLM-integrated education using RECIPE4U. EFL writing education with human-LLM collaboration holds the potential to maximize learning with minimal resources.

\paragraph{Student \& LLM with prompt recommendation}
Students can collaborate with LLMs to obtain satisfactory responses from LLM, enhancing user experiences by minimizing the effort needed to craft optimal prompts. We can categorize students' prompts based on similar intents and high satisfaction levels by combining intent detection and satisfaction estimation.
Consequently, students can have access and refer to recommended prompts with comparable intents.

\paragraph{Instructor \& LLM with learning analytics}
Instructors can gain insights into their teaching methods and contents by collaborating with LLMs. They can be aware of students’ learning objectives and questions, which is reflected in the student-ChatGPT interactions. 
Statistics and analyses of request types and occurrences initiated by students' prompts provide a detailed view of their comprehension levels related to the course material. For instance, the frequency of requests for information can enable instructors to refine and improve their teaching materials in alignment with the prevalent inquiries. 
In addition, we can also develop a misuse detection system to monitor the appropriateness of students' interactions from an educational perspective. This initiative can involve the annotation of new labels to identify inappropriate or unproductive prompts in terms of learning (e.g., asking for answers to quizzes and assignments and asking ChatGPT to generate essays by themselves).


\section{Conclusion}
In this paper, we release RECIPE4U, the first dataset on EFL learners' interaction with ChatGPT in a semester-long essay writing context. 
RECIPE4U includes 1) student-ChatGPT conversation log, 2) students' intent which is annotated with our coding schemes, 3) students' self-rated satisfaction, and 4) utterance-level essay edit history. 
Given that LLMs, including ChatGPT, are not inherently crafted for educational contexts, it is necessary to delve into the students' usage patterns of LLM in the context of EFL writing education.
We explore the EFL learners' learning process with ChatGPT in English writing education, utilizing RECIPE4U with baseline models and an in-depth analysis of students' interaction.
First, we establish baseline models for two subtasks: intent detection and satisfaction estimation.
We also analyze students' interaction through the investigation of 1) student-AI dialogue patterns, 2) statistics on essay data, and 3) students' essay edits.
We finally suggest prospective pathways for LLM-integrated education with instructor \& AI and student \& AI collaboration.

\section*{Acknowledgements}
This work was supported by Elice.
This research project has benefitted from the Microsoft Accelerate Foundation Models Research (AFMR) grant program through which leading foundation models hosted by Microsoft Azure along with access to Azure credits were provided to conduct the research.
This work was supported by Institute of Information \& communications Technology Planning \& Evaluation (IITP) grant funded by the Korea government(MSIT) (No. 2022-0-00184, Development and Study of AI Technologies to Inexpensively Conform to Evolving Policy on Ethics).


\section*{Ethics Statement}
All studies in this research project were performed under our institutional review board (IRB) approval.

There was no discrimination when recruiting and selecting EFL students and instructors regarding any demographics, including gender and age.
We set the wage per session to be above the minimum wage in the Republic of Korea in 2023 (KRW 9,260 $\approx$ USD 7.25)\thinspace\footnote{\url{https://www.minimumwage.go.kr/}}.
They were free to participate in or drop out of the experiment, and their decision did not affect the scores or the grades they received.

We deeply considered the potential risk associated with releasing a dataset containing human-written essays in terms of privacy and personal information.
We will filter out all sensitive information related to their privacy and personal information by (1) rule-based code and (2) human inspection.
To address this concern, we will run a checklist, and only the researchers or practitioners who submit the checklist can access our data.

\section*{Bibliographical References}
\vspace{-5mm}
\bibliographystyle{lrec-coling2024-natbib}
\bibliography{references/anthology, references/bibliographical}

\begin{thebibliography}{8}
\expandafter\ifx\csname natexlab\endcsname\relax\def\natexlab#1{#1}\fi

\bibitem[{Coucke et~al.(2018)Coucke, Saade, Ball, Bluche, Caulier, Leroy,
  Doumouro, Gisselbrecht, Caltagirone, Lavril, Primet, and
  Dureau}]{coucke2018snips}
Alice Coucke, Alaa Saade, Adrien Ball, Théodore Bluche, Alexandre Caulier,
  David Leroy, Clément Doumouro, Thibault Gisselbrecht, Francesco Caltagirone,
  Thibaut Lavril, Maël Primet, and Joseph Dureau. 2018.
\newblock \href {http://arxiv.org/abs/1805.10190} {Snips voice platform: an
  embedded spoken language understanding system for private-by-design voice
  interfaces}.

\bibitem[{El~Asri et~al.(2017)El~Asri, Schulz, Sharma, Zumer, Harris, Fine,
  Mehrotra, and Suleman}]{el-asri-etal-2017-frames}
Layla El~Asri, Hannes Schulz, Shikhar Sharma, Jeremie Zumer, Justin Harris,
  Emery Fine, Rahul Mehrotra, and Kaheer Suleman. 2017.
\newblock \href {https://doi.org/10.18653/v1/W17-5526} {{F}rames: a corpus for
  adding memory to goal-oriented dialogue systems}.
\newblock In \emph{Proceedings of the 18th Annual {SIG}dial Meeting on
  Discourse and Dialogue}, pages 207--219, Saarbr{\"u}cken, Germany.
  Association for Computational Linguistics.

\bibitem[{Eric et~al.(2017)Eric, Krishnan, Charette, and
  Manning}]{eric-etal-2017-key}
Mihail Eric, Lakshmi Krishnan, Francois Charette, and Christopher~D. Manning.
  2017.
\newblock \href {https://doi.org/10.18653/v1/W17-5506} {Key-value retrieval
  networks for task-oriented dialogue}.
\newblock In \emph{Proceedings of the 18th Annual {SIG}dial Meeting on
  Discourse and Dialogue}, pages 37--49, Saarbr{\"u}cken, Germany. Association
  for Computational Linguistics.

\bibitem[{Hemphill et~al.(1990)Hemphill, Godfrey, and
  Doddington}]{hemphill-etal-1990-atis}
Charles~T. Hemphill, John~J. Godfrey, and George~R. Doddington. 1990.
\newblock \href {https://aclanthology.org/H90-1021} {The {ATIS} spoken language
  systems pilot corpus}.
\newblock In \emph{Speech and Natural Language: Proceedings of a Workshop Held
  at Hidden Valley, {P}ennsylvania, June 24-27,1990}.

\bibitem[{Henderson et~al.(2014)Henderson, Thomson, and
  Williams}]{henderson-etal-2014-second}
Matthew Henderson, Blaise Thomson, and Jason~D. Williams. 2014.
\newblock \href {https://doi.org/10.3115/v1/W14-4337} {The second dialog state
  tracking challenge}.
\newblock In \emph{Proceedings of the 15th Annual Meeting of the Special
  Interest Group on Discourse and Dialogue ({SIGDIAL})}, pages 263--272,
  Philadelphia, PA, U.S.A. Association for Computational Linguistics.

\bibitem[{Shah et~al.(2018)Shah, Hakkani-Tür, Tür, Rastogi, Bapna, Nayak, and
  Heck}]{shah2018building}
Pararth Shah, Dilek Hakkani-Tür, Gokhan Tür, Abhinav Rastogi, Ankur Bapna,
  Neha Nayak, and Larry Heck. 2018.
\newblock \href {http://arxiv.org/abs/1801.04871} {Building a conversational
  agent overnight with dialogue self-play}.

\bibitem[{Zang et~al.(2020)Zang, Rastogi, Sunkara, Gupta, Zhang, and
  Chen}]{zang-etal-2020-multiwoz}
Xiaoxue Zang, Abhinav Rastogi, Srinivas Sunkara, Raghav Gupta, Jianguo Zhang,
  and Jindong Chen. 2020.
\newblock \href {https://doi.org/10.18653/v1/2020.nlp4convai-1.13}
  {{M}ulti{WOZ} 2.2 : A dialogue dataset with additional annotation corrections
  and state tracking baselines}.
\newblock In \emph{Proceedings of the 2nd Workshop on Natural Language
  Processing for Conversational AI}, pages 109--117, Online. Association for
  Computational Linguistics.

\bibitem[{Zhang et~al.(2023)Zhang, Divekar, Ubale, and
  Yu}]{zhang-etal-2023-groundialog}
Xuanming Zhang, Rahul Divekar, Rutuja Ubale, and Zhou Yu. 2023.
\newblock \href {https://doi.org/10.18653/v1/2023.bea-1.26} {{G}roun{D}ialog: A
  dataset for repair and grounding in task-oriented spoken dialogues for
  language learning}.
\newblock In \emph{Proceedings of the 18th Workshop on Innovative Use of NLP
  for Building Educational Applications (BEA 2023)}, pages 300--314, Toronto,
  Canada. Association for Computational Linguistics.

\end{thebibliography}


\begin{thebibliography}{28}
\expandafter\ifx\csname natexlab\endcsname\relax\def\natexlab#1{#1}\fi

\bibitem[{Boyd(2015)}]{boyd-2015-relations}
Maureen~P. Boyd. 2015.
\newblock \href {https://doi.org/10.1177/1086296X16632451} {{Relations Between
  Teacher Questioning and Student Talk in One Elementary ELL Classroom}}.
\newblock \emph{Journal of Literacy Research}, 47(3):370--404.

\bibitem[{Boyer et~al.(2010)Boyer, Ha, Phillips, Wallis, Vouk, and
  Lester}]{boyer-etal-2010-dialogue}
Kristy Boyer, Eun~Y. Ha, Robert Phillips, Michael Wallis, Mladen Vouk, and
  James Lester. 2010.
\newblock \href {https://aclanthology.org/W10-4356} {Dialogue act modeling in a
  complex task-oriented domain}.
\newblock In \emph{Proceedings of the {SIGDIAL} 2010 Conference}, pages
  297--305, Tokyo, Japan. Association for Computational Linguistics.

\bibitem[{Cheng(2004)}]{cheng-2004-efl}
YS~Cheng. 2004.
\newblock Efl students’ writing anxiety: Sources and implications.
\newblock \emph{English Teaching \& Learning}, 29(2):41--62.

\bibitem[{Conneau et~al.(2020)Conneau, Khandelwal, Goyal, Chaudhary, Wenzek,
  Guzm{\'a}n, Grave, Ott, Zettlemoyer, and
  Stoyanov}]{conneau-etal-2020-unsupervised}
Alexis Conneau, Kartikay Khandelwal, Naman Goyal, Vishrav Chaudhary, Guillaume
  Wenzek, Francisco Guzm{\'a}n, Edouard Grave, Myle Ott, Luke Zettlemoyer, and
  Veselin Stoyanov. 2020.
\newblock \href {https://doi.org/10.18653/v1/2020.acl-main.747} {Unsupervised
  cross-lingual representation learning at scale}.
\newblock In \emph{Proceedings of the 58th Annual Meeting of the Association
  for Computational Linguistics}, pages 8440--8451, Online. Association for
  Computational Linguistics.

\bibitem[{Demszky et~al.(2021)Demszky, Liu, Mancenido, Cohen, Hill, Jurafsky,
  and Hashimoto}]{demszky-etal-2021-measuring}
Dorottya Demszky, Jing Liu, Zid Mancenido, Julie Cohen, Heather Hill, Dan
  Jurafsky, and Tatsunori Hashimoto. 2021.
\newblock \href {https://doi.org/10.18653/v1/2021.acl-long.130} {Measuring
  conversational uptake: A case study on student-teacher interactions}.
\newblock In \emph{Proceedings of the 59th Annual Meeting of the Association
  for Computational Linguistics and the 11th International Joint Conference on
  Natural Language Processing (Volume 1: Long Papers)}, pages 1638--1653,
  Online. Association for Computational Linguistics.

\bibitem[{Devlin et~al.(2019)Devlin, Chang, Lee, and
  Toutanova}]{devlin-etal-2019-bert}
Jacob Devlin, Ming-Wei Chang, Kenton Lee, and Kristina Toutanova. 2019.
\newblock \href {https://doi.org/10.18653/v1/N19-1423} {{BERT}: Pre-training of
  deep bidirectional transformers for language understanding}.
\newblock In \emph{Proceedings of the 2019 Conference of the North {A}merican
  Chapter of the Association for Computational Linguistics: Human Language
  Technologies, Volume 1 (Long and Short Papers)}, pages 4171--4186,
  Minneapolis, Minnesota. Association for Computational Linguistics.

\bibitem[{Dowlagar and Mamidi(2023)}]{dowlagar-2023-code}
Suman Dowlagar and Radhika Mamidi. 2023.
\newblock \href {https://doi.org/10.1016/j.csl.2022.101449} {A code-mixed
  task-oriented dialog dataset for medical domain}.
\newblock \emph{Comput. Speech Lang.}, 78(C).

\bibitem[{Grassini(2023)}]{baidoo-2023-education}
Simone Grassini. 2023.
\newblock \href {https://doi.org/10.3390/educsci13070692} {{Shaping the Future
  of Education: Exploring the Potential and Consequences of AI and ChatGPT in
  Educational Settings}}.
\newblock \emph{Education Sciences}, 13(7).

\bibitem[{Ha et~al.(2012)Ha, Grafsgaard, Mitchell, Boyer, and
  Lester}]{ha-etal-2012-combining}
Eun~Young Ha, Joseph~F. Grafsgaard, Christopher Mitchell, Kristy~Elizabeth
  Boyer, and James~C. Lester. 2012.
\newblock \href {https://aclanthology.org/W12-1634} {Combining verbal and
  nonverbal features to overcome the {``}information gap{''} in task-oriented
  dialogue}.
\newblock In \emph{Proceedings of the 13th Annual Meeting of the Special
  Interest Group on Discourse and Dialogue}, pages 247--256, Seoul, South
  Korea. Association for Computational Linguistics.

\bibitem[{Han et~al.(2023{\natexlab{a}})Han, Yoo, Kim, Myung, Kim, Lim, Kim,
  Lee, Hong, Ahn, and Oh}]{han-etal-2023-recipe}
Jieun Han, Haneul Yoo, Yoonsu Kim, Junho Myung, Minsun Kim, Hyunseung Lim, Juho
  Kim, Tak~Yeon Lee, Hwajung Hong, So-Yeon Ahn, and Alice Oh.
  2023{\natexlab{a}}.
\newblock \href {https://doi.org/10.1145/3573051.3596200} {{RECIPE: How to
  Integrate ChatGPT into EFL Writing Education}}.
\newblock In \emph{Proceedings of the Tenth ACM Conference on Learning @
  Scale}, L@S '23, page 416–420, New York, NY, USA. Association for Computing
  Machinery.

\bibitem[{Han et~al.(2023{\natexlab{b}})Han, Yoo, Myung, Kim, Lim, Kim, Lee,
  Hong, Kim, Ahn, and Oh}]{han-2023-fabric}
Jieun Han, Haneul Yoo, Junho Myung, Minsun Kim, Hyunseung Lim, Yoonsu Kim,
  Tak~Yeon Lee, Hwajung Hong, Juho Kim, So-Yeon Ahn, and Alice Oh.
  2023{\natexlab{b}}.
\newblock \href {http://arxiv.org/abs/2310.05191} {Fabric: Automated scoring
  and feedback generation for essays}.

\bibitem[{Hosseini-Asl et~al.(2020)Hosseini-Asl, McCann, Wu, Yavuz, and
  Socher}]{hosseini-Asl-2020-simple}
Ehsan Hosseini-Asl, Bryan McCann, Chien-Sheng Wu, Semih Yavuz, and Richard
  Socher. 2020.
\newblock \href
  {https://proceedings.neurips.cc/paper_files/paper/2020/file/e946209592563be0f01c844ab2170f0c-Paper.pdf}
  {A simple language model for task-oriented dialogue}.
\newblock In \emph{Advances in Neural Information Processing Systems},
  volume~33, pages 20179--20191. Curran Associates, Inc.

\bibitem[{Kasneci et~al.(2023)Kasneci, Sessler, Küchemann, Bannert,
  Dementieva, Fischer, Gasser, Groh, Günnemann, Hüllermeier, Krusche,
  Kutyniok, Michaeli, Nerdel, Pfeffer, Poquet, Sailer, Schmidt, Seidel,
  Stadler, Weller, Kuhn, and Kasneci}]{kasneci-2023-chatgpt}
Enkelejda Kasneci, Kathrin Sessler, Stefan Küchemann, Maria Bannert, Daryna
  Dementieva, Frank Fischer, Urs Gasser, Georg Groh, Stephan Günnemann, Eyke
  Hüllermeier, Stephan Krusche, Gitta Kutyniok, Tilman Michaeli, Claudia
  Nerdel, Jürgen Pfeffer, Oleksandra Poquet, Michael Sailer, Albrecht Schmidt,
  Tina Seidel, Matthias Stadler, Jochen Weller, Jochen Kuhn, and Gjergji
  Kasneci. 2023.
\newblock \href {https://doi.org/https://doi.org/10.1016/j.lindif.2023.102274}
  {{ChatGPT for good? On opportunities and challenges of large language models
  for education}}.
\newblock \emph{Learning and Individual Differences}, 103:102274.

\bibitem[{Lee et~al.(2023)Lee, Myung, Han, Jin, and Oh}]{lee-2023-learning}
Changyoon Lee, Junho Myung, Jieun Han, Jiho Jin, and Alice Oh. 2023.
\newblock \href {http://arxiv.org/abs/2309.10419} {{Learning from Teaching
  Assistants to Program with Subgoals: Exploring the Potential for AI Teaching
  Assistants}}.

\bibitem[{Lin et~al.(2021)Lin, Madotto, Winata, Xu, Jiang, Hu, Shi, and
  Fung}]{lin-2021-bitod}
Zhaojiang Lin, Andrea Madotto, Genta Winata, Peng Xu, Feijun Jiang, Yuxiang Hu,
  Chen Shi, and Pascale~N Fung. 2021.
\newblock \href
  {https://datasets-benchmarks-proceedings.neurips.cc/paper_files/paper/2021/file/6364d3f0f495b6ab9dcf8d3b5c6e0b01-Paper-round1.pdf}
  {Bitod: A bilingual multi-domain dataset for task-oriented dialogue
  modeling}.
\newblock In \emph{Proceedings of the Neural Information Processing Systems
  Track on Datasets and Benchmarks}, volume~1. Curran.

\bibitem[{Lu et~al.(2023)Lu, Fan, Houghton, Wang, and
  Wang}]{lu-2023-readingquizmaker}
Xinyi Lu, Simin Fan, Jessica Houghton, Lu~Wang, and Xu~Wang. 2023.
\newblock \href {https://doi.org/10.1145/3544548.3580957} {{ReadingQuizMaker: A
  Human-NLP Collaborative System That Supports Instructors to Design
  High-Quality Reading Quiz Questions}}.
\newblock In \emph{Proceedings of the 2023 CHI Conference on Human Factors in
  Computing Systems}, CHI '23, New York, NY, USA. Association for Computing
  Machinery.

\bibitem[{Marineau et~al.(2000)Marineau, Wiemer-Hastings, Harter, Olde,
  Chipman, Karnavat, Pomeroy, Rajan, Graesser, Group
  et~al.}]{marineau-2000-classification}
Johanna Marineau, Peter Wiemer-Hastings, Derek Harter, Brent Olde, Patrick
  Chipman, Ashish Karnavat, Victoria Pomeroy, Sonya Rajan, Art Graesser,
  Tutoring~Research Group, et~al. 2000.
\newblock Classification of speech acts in tutorial dialog.
\newblock In \emph{Proceedings of the Workshop on Modeling Human Teaching
  Tactics and Strategies of ITS 2000}, pages 65--71.

\bibitem[{Markel et~al.(2023)Markel, Opferman, Landay, and
  Piech}]{markel-2023-gpteach}
Julia~M. Markel, Steven~G. Opferman, James~A. Landay, and Chris Piech. 2023.
\newblock \href {https://doi.org/10.1145/3573051.3593393} {{GPTeach:
  Interactive TA Training with GPT-Based Students}}.
\newblock In \emph{Proceedings of the Tenth ACM Conference on Learning @
  Scale}, L@S '23, page 226–236, New York, NY, USA. Association for Computing
  Machinery.

\bibitem[{Mercer(2008)}]{mercer-2008-seeds}
Neil Mercer. 2008.
\newblock \href {https://doi.org/10.1080/10508400701793182} {The seeds of time:
  Why classroom dialogue needs a temporal analysis}.
\newblock \emph{Journal of the Learning Sciences}, 17(1):33--59.

\bibitem[{Ozkose-Biyik and Meskill(2015)}]{ozkose-2015-plays}
Cagri Ozkose-Biyik and Carla Meskill. 2015.
\newblock \href {http://www.jstor.org/stable/43893787} {{Plays Well With
  Others: A Study of EFL Learner Reciprocity in Action}}.
\newblock \emph{TESOL Quarterly}, 49(4):787--813.

\bibitem[{Pelau et~al.(2021)Pelau, Dabija, and Ene}]{pelau-2021-makes}
Corina Pelau, Dan-Cristian Dabija, and Irina Ene. 2021.
\newblock \href {https://doi.org/https://doi.org/10.1016/j.chb.2021.106855}
  {{What makes an AI device human-like? The role of interaction quality,
  empathy and perceived psychological anthropomorphic characteristics in the
  acceptance of artificial intelligence in the service industry}}.
\newblock \emph{Computers in Human Behavior}, 122:106855.

\bibitem[{Qadir(2023)}]{qadir-2023-engineering}
Junaid Qadir. 2023.
\newblock \href {https://doi.org/10.1109/EDUCON54358.2023.10125121}
  {{Engineering Education in the Era of ChatGPT: Promise and Pitfalls of
  Generative AI for Education}}.
\newblock In \emph{2023 IEEE Global Engineering Education Conference (EDUCON)},
  pages 1--9.

\bibitem[{Rasor et~al.(2011)Rasor, Olney, and D'Mello}]{rasor-2011-student}
Travis Rasor, Andrew Olney, and Sidney D'Mello. 2011.
\newblock Student speech act classification using machine learning.
\newblock In \emph{Proceedings of the Twenty-Fourth International Florida
  Artificial Intelligence Research Society Conference}.

\bibitem[{Schuster et~al.(2019)Schuster, Gupta, Shah, and
  Lewis}]{schuster-etal-2019-cross-lingual}
Sebastian Schuster, Sonal Gupta, Rushin Shah, and Mike Lewis. 2019.
\newblock \href {https://doi.org/10.18653/v1/N19-1380} {Cross-lingual transfer
  learning for multilingual task oriented dialog}.
\newblock In \emph{Proceedings of the 2019 Conference of the North {A}merican
  Chapter of the Association for Computational Linguistics: Human Language
  Technologies, Volume 1 (Long and Short Papers)}, pages 3795--3805,
  Minneapolis, Minnesota. Association for Computational Linguistics.

\bibitem[{Silaban and Marpaung(2020)}]{silaban-2020-analysis}
Suardani Silaban and Tiarma~Intan Marpaung. 2020.
\newblock An analysis of code-mixing and code-switching used by indonesia
  lawyers club on tv one.
\newblock \emph{Journal of English Teaching as a Foreign Language}, 6(3):1--17.

\bibitem[{Sun and Fan(2022)}]{sun-2022-effects}
Bo~Sun and Tingting Fan. 2022.
\newblock The effects of an awe-aided assessment approach on business english
  writing performance and writing anxiety: A contextual consideration.
\newblock \emph{Studies in Educational Evaluation}, 72:101123.

\bibitem[{Trust et~al.(2023)Trust, Whalen, and Mouza}]{whalen-2023-chatgpt}
Torrey Trust, Jeromie Whalen, and Chrystalla Mouza. 2023.
\newblock \href {https://www.learntechlib.org/p/222408} {{Editorial: ChatGPT:
  Challenges, Opportunities, and Implications for Teacher Education}}.
\newblock \emph{Contemporary Issues in Technology and Teacher Education},
  23(1):1--23.

\bibitem[{Wen et~al.(2020)Wen, Lu, and Reddy}]{wen-etal-2020-medal}
Zhi Wen, Xing~Han Lu, and Siva Reddy. 2020.
\newblock \href {https://doi.org/10.18653/v1/2020.clinicalnlp-1.15} {{M}e{DAL}:
  Medical abbreviation disambiguation dataset for natural language
  understanding pretraining}.
\newblock In \emph{Proceedings of the 3rd Clinical Natural Language Processing
  Workshop}, pages 130--135, Online. Association for Computational Linguistics.

\end{thebibliography}

\section*{Language Resource References}
\vspace{-5mm}
\bibliographystylelanguageresource{lrec-coling2024-natbib}
\bibliographylanguageresource{references/anthology, references/language_resources}


\end{document}